\documentclass[journal]{IEEEtran}
\ifCLASSINFOpdf
\else
\fi
\usepackage{amsmath}
\usepackage{cite}
\usepackage{graphicx}
\usepackage{color}
\usepackage[center]{caption}
\usepackage{algorithm}
\usepackage{algorithmic}
\begin{document}
%
\title{Call Detail Records Driven Anomaly Detection and Traffic Prediction in Mobile Cellular Networks}
%
%
%

\author{Kashif~Sultan,
        Hazrat~Ali,
				Zhongshan~Zhang*
\thanks{Kashif Sultan and Zhongshan Zhang are with School of Computer and Communication Engineering, University of Science and Technology Beijing, 30 Xueyuan Road, Haidian District, Beijing, 100083, PR China. e-mail: kashif\_rao@outlook.com; kashif\_sultan@xs.ustb.edu.cn; zhangzs@ustb.edu.cn}
\thanks{Hazrat Ali is with Department of Electrical Engineering, COMSATS University Islamabad, Abbottabad Campus, 22060, Abbottabad, Pakistan. e-mail: hazratali@ciit.net.pk. ORCID: https://orcid.org/0000-0003-3058-5794}
\thanks{Manuscript received February 2018; revised xx xx, 2018.}
\thanks{This work was supported by the key project of the National Natural Science Foundation of China (No. 61431001), Beijing Natural Science Foundation (L172026), Key Laboratory of Cognitive Radio and Information Processing, Ministry of Education (Guilin University of Electronic Technology), and the Foundation of Beijing Engineering and Technology Center for Convergence Networks and Ubiquitous Services (*Corresponding author: Zhongshan Zhang)}
}
%
%

\markboth{IEEE Access Journal,~Vol.~xx, No.~xx, January~20xx}%
{Kashif \MakeLowercase{\textit{et al.}}: CDR  Data Analytics for Anomaly Detection and Prediction}
%



\maketitle

\begin{abstract}
Mobile networks possess information about the users as well as the network. Such information is useful for making the network end-to-end visible and intelligent. Big data analytics can efficiently analyze user and network information, unearth meaningful insights with the help of machine learning tools. Utilizing big data analytics and machine learning, this work contributes in three ways. First, we utilize the \emph{call detail records} (CDR) data to detect anomalies in the network. For authentication and verification of anomalies, we use k-means clustering, an unsupervised machine learning algorithm. Through effective detection of anomalies, we can proceed to suitable design for resource distribution as well as fault detection and avoidance. Second, we prepare anomaly-free data by removing anomalous activities and train a neural network model. By passing anomaly and anomaly-free data through this model, we observe the effect of anomalous activities in training of the model and also observe mean square error of anomaly and anomaly free data. Lastly, we use an autoregressive integrated moving average (ARIMA) model to predict future traffic for a user. Through simple visualization, we show that anomaly free data better generalizes the learning models  and performs better on prediction task. 
\end{abstract}


%
\IEEEpeerreviewmaketitle

\section{Introduction}
%
%
%
%
\IEEEPARstart{M}{obile} technologies and cellular networks are getting smarter day by day, so mobile phone devices such as smartphones, tablets, wearable devices as well as mobile phone subscribers are increasing rapidly. According to report presented by Ericsson, the mobile devices have surpassed the world population \cite{devices}. Due to such a huge growth in mobile devices and mobile phone subscribers, the congestion of mobile network is not unusual. Hence the provision of best quality of services for such a huge number of mobile phone subscribers is challenging. With the massive growth of mobile devices and mobile phone subscribers, the data generated from these devices is increasing explosively. According to CISCO survey, the data increased 4000-fold during the last ten years \cite{Cisco}. From CISCO report, global mobile data traffic is generating 24 Exabyte (EB) data per month and this trend is continuously rising \cite{Cisco}. This huge data has following 4Vs characteristics making it different from the traditional data.
 %
\\\emph{Volume}: The data has very large volume; of order of picobytes (PB). As reported in \cite{rupanagunta2012mine}, 12 Terabytes (TB) data is generated by Twitter every day. On average, 1.2 Zeta bytes of data is being produced every year since 2012 and this value is continuously rising \cite{callerkey, Infographic}.
\\\emph{Velocity}: The flow rate of data at which the data goes in or out from mobile devices and mobile network is termed velocity of data.  This determines the dynamic nature of the data and big data is highly dynamic. 
\\\emph{Variety}: This huge and dynamic data comes from various sources and occurs in different formats such as structured, unstructured and semi-structured.
\\\emph{Value}: According to IDC report, a very famous group for big data research activities, big data technologies describes a new generation of technology, designed to extract value from a huge volume of a wide variety of data \cite{gantz2011extracting}.

\begin{figure}[ht!]
\centering
\vspace{0.2cm}
\graphicspath{{./figures/}}
\includegraphics[scale = 0.3]{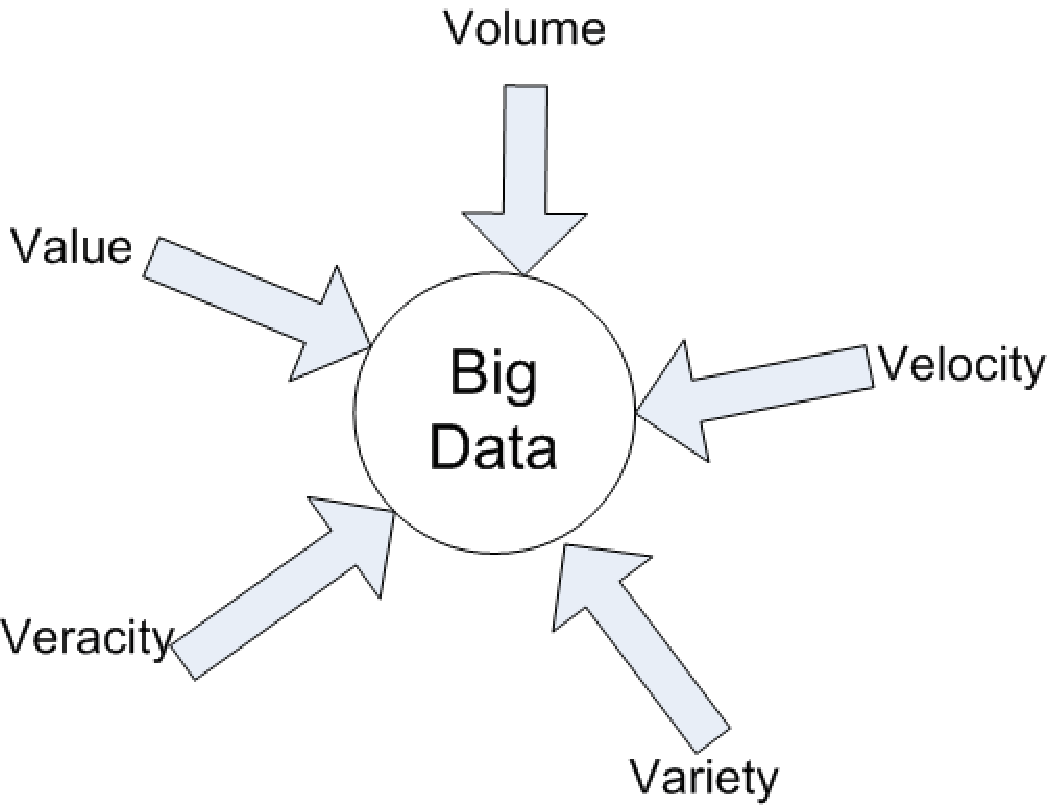}
\caption{Attributes of big data}
\label{fig:bigdata}
\end{figure}

From the above discussions, data with 4Vs characteristics (\emph{volume, velocity, variety and value}) is known as big data. The 4Vs characteristics of big data is shown in Figure \ref{fig:bigdata}. Hence, the management and analysis of such big data is termed as big data analytics. Big data analytics is a broader term and encompasses hardware and software solutions for efficient analysis and management of the data. In big data analytics, statistical machine learning and algorithms are applied on vast amount of complex data, which unveil hidden information from the data \cite{king2013big}. Big data analytics is different from the traditional data analytics. As in traditional data analytics, we choose only random samples from the entire data set and apply data analytics on selected partial dataset. Decisions are made on the basis of data analysis of partial dataset. Hence results obtained from this partial data analytics are not essentially accurate and precise. Not long ago, big data anlaytics and machine learning tools were not developed enough to analyze entire dataset, attributed to limitations on available computational resources. Recently, with advancements in such technologies, the analytics of big datasets has become feasible. Hence, the results generated by big data analytics are more precise and accurate as compared to traditional analytics. For network data analytics applications such as network optimization, resource allocation, identification of sleeping cells and proactive caching, we need accurate and precise information, which can be obtained through big data analytics. 

With emergence of 5G networks, there will be hundreds of billions of connected devices \cite{andrews2014will} and comparable number of network nodes. Hence, by analyzing information such as \emph{call details record} (CDR) information, \emph{reference signal receive power} (RSRP), and \emph{location information} from core network and radio access nodes, one can improve the network performance as well as give the best quality of service (QoS) and quality of experience (QoE) to mobile phone subscribers. 

In this work, we exploit the CDR information of mobile networks. This CDR information helps in determining the user activities at a particular date and time. In this work, we name the unexpected or abnormal behavior of the user as an \emph{anomaly}. Such type of unusual activity degrades the network performance. Anomalies in the network may occur due to different issues such as network failure, sleeping cells, overloading of traffic and low coverage area. Hence, successful anomaly detection and anomaly removal are useful in optimizing network performance. We will discuss this in detail in subsequent section. This paper contributes as below:
\begin{description}
\item[$\bullet$] We present a concept of identifying anomalies in CDR data and verify the anomalies through clustering algorithm. We then provide an insight into how the anomalous data can be removed to obtain anomaly free data. 
 \item[$\bullet$] We present an artificial neural network model trained with anomalous as well as anomaly free data. Through mean square error calculation, we show that the neural network model learns better information from the data if the anomalies are removed.
  \item[$\bullet$] We also explore the use of ARIMA model for future activity prediction of mobile subscribers.
\end{description}

The rest of the paper is organized as follows: In Section  \ref{sec:RelatedWork} we discuss related work. In Section \ref{sec:CDR} we describe importance and use case of CDR data analytics in optimizing network performance and improving the QoS for mobile subscribers. In Section \ref{sec:SystemModel}, we present the system model and also describe the dataset used. We present anomaly detection by ground truth data and machine learning algorithms and discuss the verification in Section \ref{sec:MachineLearning}. In Section \ref{sec:PredictionModels}, we present neural network based training model for comparing mean square error of anomaly and anomaly-free data. In Section \ref{sec:ARIMA} we present ARIMA model for predicting future activities of the user. Finally, we conclude our work in Section \ref{sec:Conclusion}.


\section{Related Work}
\label{sec:RelatedWork}
Anomaly detection has been reported before in work as summarized below: 

Naboulsi et al., \cite{naboulsi2014classifying} presented a framework, in which a large scale CDR dataset was sub-categorized according to the history of activities. The framework reported in \cite{naboulsi2014classifying} determines the irregular and unexpected activities termed as anomalies. The authors in \cite{soto2011automated, zoha2014solution} used k-means clustering techniques for determining regions of interest such as commercial areas, residential areas, office areas, and recreational areas etc.
The authors in \cite{munz2007traffic, lima2010anomaly} also used k-means clustering for anomaly detection purpose. The authors divided data into clusters of anomalous data and normal data.
The authors in \cite{karatepe2014anomaly} analyzed CDR information of wireless network and detected anomalies by rule based approach.
Authors in \cite{gething2011can, bengtsson2011improved} presented the significance of CDR data analysis in case of natural disasters.

The CDRs being generated daily are huge in number. The CDR data contains valuable insights that can be used for the benefit of the network operators as well as subscribers. A milestone for such a huge CDRs data analysis is presented in \cite{bouillet2012experience}. The authors in \cite{bouillet2012experience} have presented a stream processing model which is able to analyze 6 billion CDRs generated per day. The model has the ability to support higher throughput, lower latency and fault tolerance. Parallel processing, de-duplication and easy to use platform for network operators are the main aspects of the model.


The authors in \cite{zheng2016big} presented a big data analytics based model for optimizing 5G networks and showed that the network will be faster and proactive with the aid of big data analytics. The authors in \cite{fan2016dynamic} showed that bandwidth (resource) can be efficiently distributed with the aid of big data analytics. The authors in \cite{imran2014challenges} presented that self-organizing network (SON) which will be used for enabling 5G, can be efficiently implemented with big data analytics. Such a framework based on SON was named a BSON \cite{imran2014challenges}.
The authors in \cite{bacstuug2015big} showed that big data analytics will be helpful for proactive caching which is very important for empowering 5G.

Motivated from the literature, we use k-means clustering algorithm and detect the anomalous behavior of the users. Our work is different from the previously reported work which was limited to anomaly detection only. We perform verification of the anomaly detection through comparison with ground truth data. After successful anomaly detection and verification, we prepare anomaly-free data. We also train a neural network model for observing mean square error of anomaly and anomaly-free data. Finally, we train ARIMA prediction model for predicting users' future activities. Furthermore, we discuss that such type of insights are also helpful towards 5G networks' requirements such as proactive caching, maximum throughput and close-to-zero latency.

\section{The Importance of CDR data and Use Case}
\label{sec:CDR}
CDR data contains information about a subscriber's phone usage. The CDR data includes identification code of caller/receiver, the location of base tower station (BTS), direction of the voice call and activity type (i.e., SMS or call). The CDR data of a cellular network can be used for analyzing user or network behavior. With the aid of CDR data analysis, one may extract information and detect unusual events of critical significance e.g., a terrorism activity, earthquakes, floods, Christmas eve, soccer world cup, black Friday etc. If there is any such type of natural or unnatural event, the mobile subscriber activities (CDR activities) will be increased (or may be decreased). The CDR activity can also be used to detect the movements of mobile phone subscriber i.e., the flow of humans during a natural disaster occurrence.
If CDR activities at particular time and location are dramatically increasing, the rise in activity level can be linked with events such as a road accident on a highway, an usual blast or terror activity in a mall, market, earth quake or other natural disaster.  Under such circumstances, people start calling, texting to their relatives or friends or rescue services resulting in exponential rise in the CDR activity. People also tend to move away from such a location directly showing a rise in users' movement away from a particular location \cite{gething2011can, bengtsson2011improved}. Thus, CDR data analysis provides spatio-temporal information of mobile users, which means the exact location and time of a particular event can be determined. If there is flood, earthquake or any terrorism activity in a particular region, the CDR activities (calls, SMSs) will be increased in that region as mobile phone subscribers will give calls, send SMS to other people to inform about such activity in that region and will ask for rescue and help. Besides, there will be tendency of movements from the effected location to safer venues.

If there is information available about the CDR activities of a cellular network and a massive increase occurs in CDR activities at particular day or specific time frame then unusual events can be anticipated. In the event of unusual circumstances, provision of rescue services can be accelerated and help activity can be enhanced. Thus, precious lives could be saved.

Because of such type of abnormal behavior or anomalous activities, the performance of network will be down and mobile phones subscribers will have poor QoS. However, after detecting and predicting such type of anomalous activities for the next time frame, the network operator can provide some extra resources for specific time frame on a particular area. In this way the mobile phone subscribers can have improved QoS. These extra resource allocation is in the form of allocating more bandwidth to a particular area, increasing network coverage area by sending small cell nodes as drones. Hence, network congestion can be avoided and a better QoS can be ensured.
Beside event detection, the analysis of CDR activity can also be used to enhance the QoS and QoE and of mobile phone subscribers. As CDR dataset has information about the number of voice calls, duration of voice calls, caller ID, number of SMS, anlaysis of bigger CDR dataset for records spanning over several months/years is possible through machine learning and big data analytics. We can thus predict mobile subscribers' behavior i.e., the number of voice calls or the duration of voice calls for next time slot.

\section{System Model and Dataset Description}
\label{sec:SystemModel}
\begin{figure}[tp!]
\centering
\vspace{0.2cm}
\graphicspath{{./figures/}}
\includegraphics[width=2.5in,height=3in]{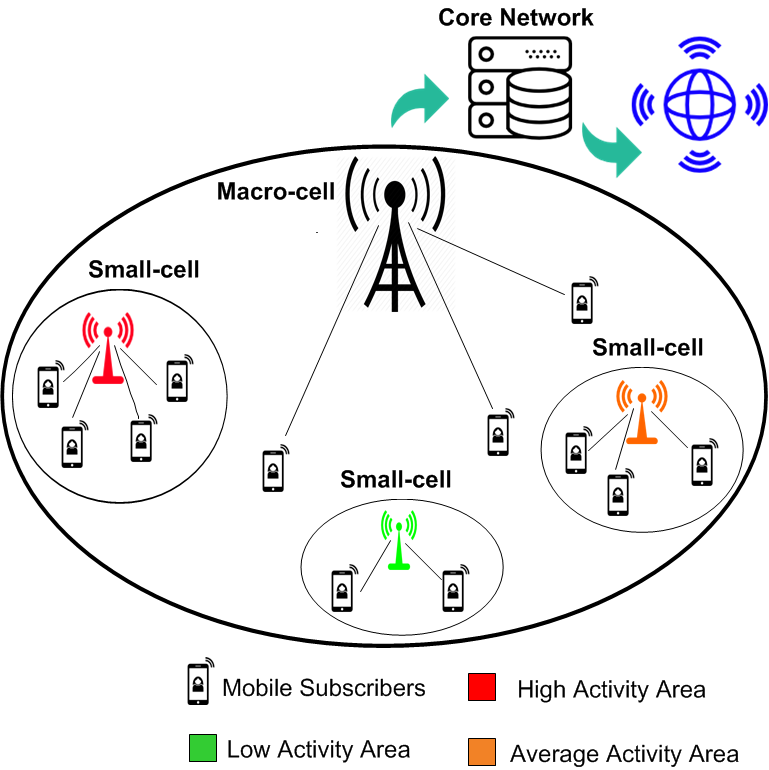}
\caption{System Model}
\label{fig:System}
\end{figure}

\begin{figure}[h]
\centering
\vspace{0.1cm}
\graphicspath{{./figures/}}
\includegraphics[width=2.8in]{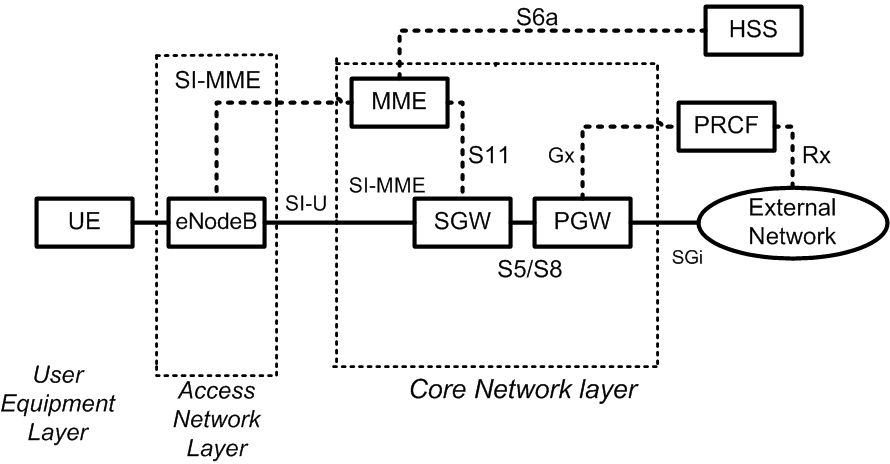}
\caption{Network Architecture}
\label{fig:Network}
\end{figure}

A generic architecture of mobile communication consists of three components; (i) mobile users or clients (ii) middle tier or access networks and (iii) back-end system or core network. Mobile users are in small cells as well as in macro-cells. Macro-cells are connected with core network and vice versa. Due to densification in next generation networks (billions of connected devices), there will be huge number of small-cells as well as macro-cells \cite{andrews2014will}. Typically, mobile users' activities are different in different cells at a particular time and area. With these considerations, we make our system model and divide the middle tier of the generic architecture according to mobile users' activity level. The hypothetical representation in Figure \ref{fig:System} shows that there are three types of areas, low activity area, average activity area and high activity area. The mobile resource requirements are different in these areas. We analyze CDR data of mobile users for observing and predicting the activities of the mobile users.
We utilize CDR data from cellular network, as available to us from  \cite{ctu-personal-20120315,mcdiarmid2013nodobo,WinNT}. As discussed above, the architecture for the network consists of three layers. The layer one consists of users' mobile devices or users' equipment, layer two is used to connect users' mobile devices or users' equipment via evolved nodeB (eNodeB). The third layer consists of logical nodes such as mobility management entity (MME), packet data network gateway (PGW) and service gateway (SGW), etc. The CDR information we use is collected from layer 3 of the network. This CDR data information is used to observe the behavior of \emph{the users} as well as \emph{the network}. From this CDR information we determine anomaly and its whereabouts in the network. The network model is shown in Figure \ref{fig:Network}.

\subsection{Description of Dataset}
We use three different datasets of cellular networks. The first CDR dataset is obtained from CRAWDAD community. CRAWDAD (\emph{Community Resource for Archiving Wireless Data Dartmouth}) is a wireless network data resource for the research community \cite{ctu-personal-20120315}. Table \ref{tab:table1} shows the CRAWDAD dataset and its associated fields. This dataset contains the mobile phone records of 142 days from September 2010 to February 2011. The dataset has following fields:\\
\textbf{Date:} The date of the users' activity are represented as YYYYMMDD. Users' activity is in the form of incoming call, outgoing call, incoming SMS and outgoing SMS.\\
\textbf{Time:} Time of the users' activity which is hhmmss.\\
\textbf{Type:} Type of the users' activity (voice call, SMS).\\
\textbf{Direction:} direction of the users' activity (incoming or outgoing).\\
\textbf{Duration:} Duration of the voice call.

\begin{table}[h]
\renewcommand{\arraystretch}{1.7}
\caption{CRAWDED dataset fields}
\label{tab:table1}
\centering
\begin{tabular}{|l|p{1cm}|p{1cm}|p{2cm}|p{1.3cm}|}
\hline
\textbf{Date} & \textbf{Time} & \textbf{Type} & \textbf{Direction} & \textbf{Duration[s]} \\
\hline
20100916 & 130748 & Voice & Incoming & 18\\ \hline
20100916 & 133131 & Voice & Outgoing & 99 \\ \hline
20100916 & 131324 & Voice & Incoming & 214 \\ \hline
20100916 & 131735 & Voice & Incoming & 72\\ \hline
20100916 & 135342 & Voice & Incoming & 37 \\ \hline
\end{tabular}
\end{table}

The second CDR dataset is obtained from Nodobo, which is a suite of software developed at the \emph{University of Strathclyde}, and allows precise capture and replay of smartphone user interactions sessions  \cite{nodobo,mcdiarmid2013nodobo}. This dataset was collected during a study of mobile phone usage of 27 high school students. The datset contains the mobile records for a period of six months from September 2010 to February 2011 \cite{mcdiarmid2013nodobo}. Table \ref{tab:table2} shows the second dataset and its associated fields. This dataset includes 13035 voice call records, 83542 messages records and other related data. The dataset has the following fields:\\
\textbf{User:} This field contains the ID of the subscriber or caller.\\
\textbf{Other:} This contains ID of the receiver.\\
\textbf{Direction:} Direction of the user's activity (incoming or outgoing).\\
\textbf{Duration:} Duration of the voice calls.\\
\textbf{Timestamp:} This field contains date and time of the user's activity.\\

\begin{table}[h]
\renewcommand{\arraystretch}{1.7}
\caption{Nodobo dataset fields}
\label{tab:table2}
\centering
\begin{tabular}{|l|p{1cm}|p{1cm}|p{1cm}|p{2cm}|}
\hline

\textbf{User} & \textbf{Other} & \textbf{Direction} & \textbf{Duration} & \textbf{Timestamp} \\
\hline
7641036117 & 7588304495 & Incoming & 1224 & Thu Sep 9 19:35:37 100 2010\\ \hline
7981267897 & 7784425582 & Outgoing & 474 & Thu Sep 9 18:43:44 100 2010 \\ \hline
7981267897 & 7743039441 & Missed & 0 & Thu Sep 9 19:51:30 100 2010 \\ \hline
7981267897 & 7784425582 & Outgoing & 0 & Thu Sep 9 20:57:55 100 2010\\ \hline
7981267897 & 7784425582 & Outgoing & 605& Fri Sep 10 20:17:00 100 2010 \\ \hline
\end{tabular}
\end{table}

The third dataset is obtained from open big data database of Dandelion API web forum. The dataset which we use from this open big data database is for \emph{Telecom Italia}. This CDR available from the Telecom Italia is for the city of Milano. The dataset includes CDRs for a period of two months from Ist November 2013 to Ist January 2014. Table \ref{tab:table3} represents sample of this dataset. This dataset provides the following information:\\
\textbf{Grid ID:} This field contains the identification number of the grid.\\
\textbf{Time stamp:} The time of the activity (in milisecond).\\
\textbf{Received SMS:} It represents number of SMS received at each time step.\\
\textbf{Sent SMS:} It represents number of SMS sent at each time step.\\
\textbf{Incoming Calls:} Duration of the incoming calls.\\
\textbf{Outgoing Calls:} Duration of outgoing calls.\\

\begin{table}[h]
\renewcommand{\arraystretch}{1.7}
\caption{Telecom Italia dataset fields}
\label{tab:table3}
\centering
\begin{tabular}{|l|p{1cm}|p{1cm}|p{1cm}|p{1cm}|p{1cm}|}
\hline

\textbf{Grid ID} & \textbf{Timestamp} & \textbf{Recevied SMS activity}\ & \textbf{Sent SMS activity}\ & \textbf{Incoming Calls activity} & \textbf{Outgoing Calls activity}  \\
\hline
1 & 10 &0.2724 & 0.1127  & 0.0035 & 0.0807\\ \hline
10 & 20 & 0.0101 & 0.0693 & 0.0573 & 0.0446\\ \hline

\end{tabular}
\end{table}

\subsection{Data Pre-processing}
The data from these sources contain missing entries or noise causing misleading pattern. Data pre-processing step removes such type of irregularities. Through pre-processing, we obtain data now ready for further analysis. 

\section{Anomaly Detection and Verification}
\label{sec:MachineLearning}
In this section, we discuss anomaly detection with machine learning tools. We then verify anomalies through visualization of the ground truth data and comparison with the anomalies detected through clustering. 
\subsection{k-means Clustering}
Clustering is a process of partitioning a group of data into small number of clusters and sub-groups. In k-means clustering, $'n'$ is number of objects or data points to be portioned and $'k'$ is the number of clusters or sub-groups. The k-means clustering algorithm is summarized in Algorithm \ref{algo:algo1}.

%
\begin{algorithm}
\caption{Algorithm}
\label{algo:algo1}
\begin{algorithmic}[H]

\STATE 
$input \gets$ K(number of cluster)
\STATE
$input \gets$ training set ${x^{1}, x^{2},........,x^{m}}$

$x^{i} \in \Re^n$ (drop $x_{0}$=1 convention )

\STATE 
Randomly initialize $K$ cluster centroids \\
$u_{1},u_{2},......,u_{K} \in \Re^n$

\REPEAT
\FOR{i =1 \TO m}
        \STATE $c_{i}$ :=index(from 1 to K) of cluster centroid closed to $x_{i}$
      \ENDFOR
\FOR{k =1 to K}
        \STATE $u_{k}$ := average (mean) of points assigned to cluster k
      \ENDFOR
\UNTIL (converge)


\end{algorithmic}
\end{algorithm}
After k-means clustering, we obtain number of clusters with different data points. Thus, the data is divided into different clusters. As one can anticipate, cluster of fewer objects or data points is the cluster of anomalous activities. 

\subsection{Anomaly Detection through Ground truth Data}
\begin{figure}[hbt]
\centering
\vspace{0.3cm}
\graphicspath{{./figures/}}
\includegraphics[width=2.5in]{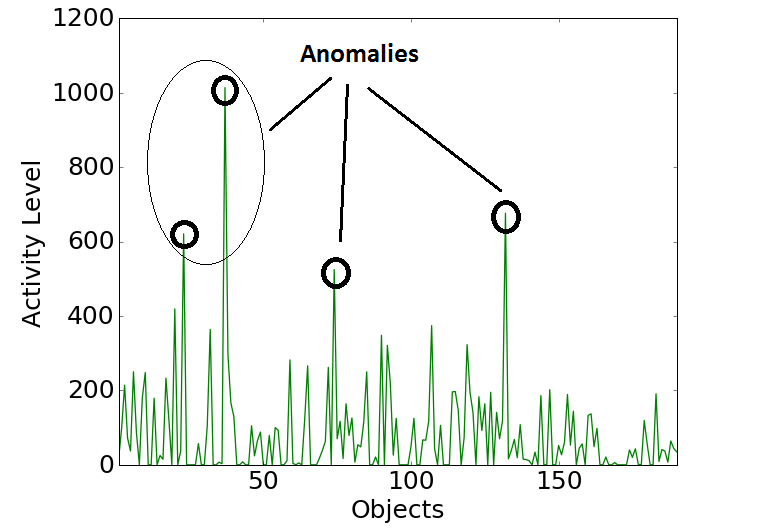}
\caption{Anomaly detection with ground truth data (first dataset)}
\label{fig:Anomaly}
\end{figure}

\begin{figure}[hbt]
\centering
\vspace{0.3cm}
\graphicspath{{./figures/}}
\includegraphics[width=2.5in]{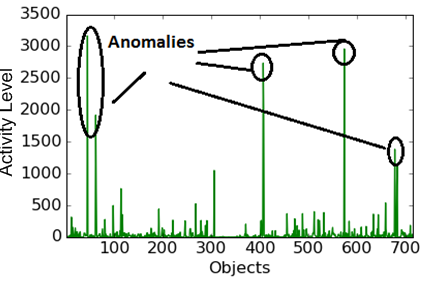}
\caption{Anomaly detection with ground truth data (second dataset)}
\label{fig:Anomaly3}
\end{figure}

\begin{figure}[hbt]
\centering
\vspace{0.3cm}
\graphicspath{{./figures/}}
\includegraphics[width=2.5in]{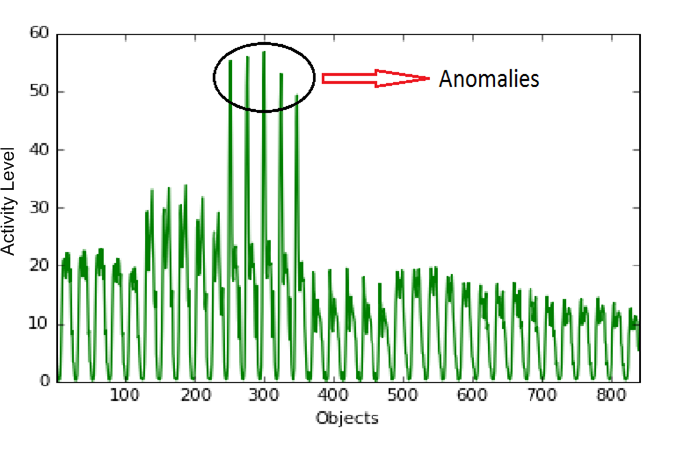}
\caption{Anomaly detection with ground truth data (third dataset)}
\label{fig:Anomaly2}
\end{figure}

Figure \ref{fig:Anomaly} shows the activity of users over a week from September 16 to September 24 in the year of 2010. It is observed from Figure \ref{fig:Anomaly} that the activity level is very high in some parts of entire week’s activity. These higher levels are showing an unexpected and irregular behavior of the users, so we consider these activities as anomalies. It is also observed from entries of the dataset that these anomalies occurred on September 17 (5.32 pm), September 19 (5.40 pm), September 22 (12.48 pm) and September 24 (9.37 pm). It can also be observed from the plot of ground truth data and entries of the dataset that on September 24 during 5pm to 6pm, there was zero activity. These activities are also considered as anomalies because at this time period as zero activity might have been due to a network failure. Similarly, the users' activities for second and third datasets are shown in Figure \ref{fig:Anomaly3} and Figure \ref{fig:Anomaly2}, respectively. Anomalies in second dataset are also obvious from Figure \ref{fig:Anomaly2}. These anomalous activities are also helpful to identify region of interest (ROI) e.g., locations with high density of users such as shopping malls, hospitals, or stadium; or identify events of interest e.g., natural disaster, fatal road accidents, or terror activity. 

\subsection{Anomaly Detection through k-means}
\begin{figure}[hbt]
\centering
\vspace{0.3cm}
\graphicspath{{./figures/}}
\includegraphics[width=2.5in]{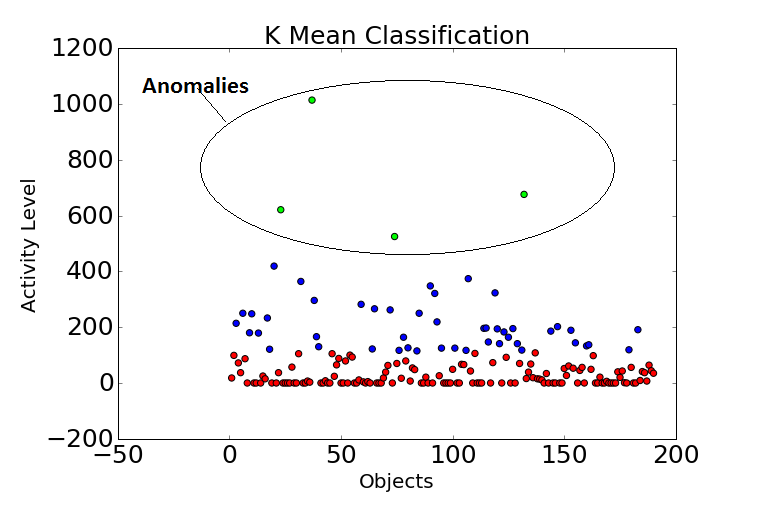}
\caption{Anomaly detection through k-means clustering (first dataset)}
\label{fig:kmean}
\end{figure}

\begin{figure}[hbt]
\centering
\vspace{0.3cm}
\graphicspath{{./figures/}}
\includegraphics[width=2.5in]{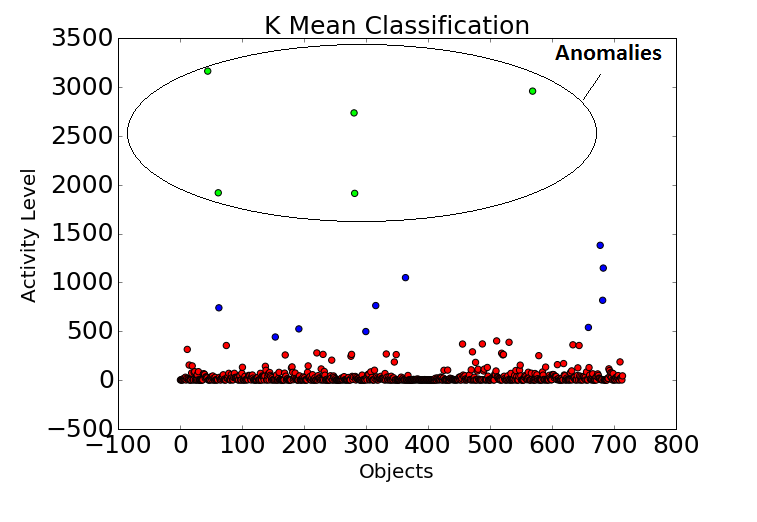}
\caption{Anomaly detection through k-means clustering (second dataset)}
\label{fig:kmean2}
\end{figure}

As anomalous activities are unexpected and irregular making them unique and fewer in number than normal activities. Through k-means clustering algorithms, normal activities are grouped into the same clusters different than the one in which abnormal activities are placed. As shown in Figure \ref{fig:kmean}, there are three clusters. Objects or users which lie in the range of activity level 1-150 are grouped in cluster 1, users which lie in the range of activity level 150-400 are grouped in cluster 2 and remaining users with much higher activity level are grouped in cluster 3. From figure \ref{fig:kmean}, we observe that most of the users are in cluster 1 and cluster 2. On the other hand, cluster 3 has fewer users with higher activity level. Cluster 3 represents abnormal activities of the users and hence characterized as cluster of anomalous activities. It is verified from Figure \ref{fig:Anomaly} and Figure \ref{fig:kmean} that k-means clustering algorithm detects a similar pattern of anomalous activities as seen from the plot of ground truth data. Similarly, Figure \ref{fig:Anomaly2} and Figure \ref{fig:kmean2} verify the anomalies for the second dataset.

As in case of future generation network or 5G, which will be proactive, have close to zero latency, maximum throughput and connected with billions of devices \cite{andrews2014will}, the exact prediction of anomalies will perform a key role in fulfilling requirements of these networks. For implementation of such ultra fast hyper dense networks (5G), there must be intelligent prediction model that could forecast the future activities. The fuel of such model is precise and accurate information. For example if prediction model is not trained with clean and anomaly-free data, the model would fail to predict. In the next section, we present an approach to prepare anomaly-free data and pass it through neural network based training model and ARIMA time series prediction model for observing error difference and predicting future 
activities.

\begin{figure}[hbt]
\centering
\vspace{0.3cm}
\graphicspath{{./figures/}}
\includegraphics[width=2.5in]{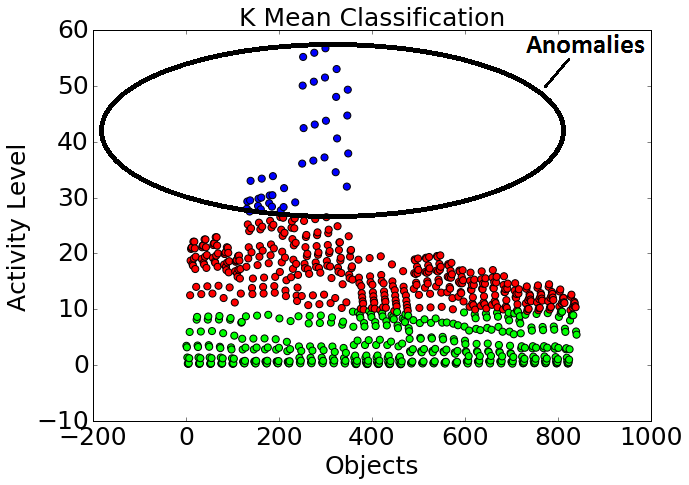}
\caption{Anomaly detection through k-means clustering (third dataset)}
\label{fig:kmean3}
\end{figure}

\section{Preparation of anomaly-free data and Mean Square Error Evaluation}
\label{sec:PredictionModels}

\subsection{Preparation of anomaly-free data}
\begin{figure}[hbt]
\centering
\vspace{0.3cm}
\graphicspath{{./figures/}}
\includegraphics[width=2.5in]{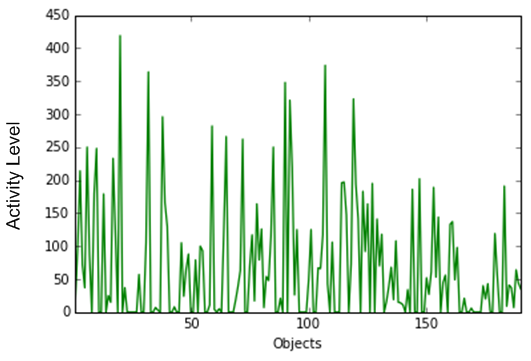}
\caption{Anomaly-free data (first dataset)}
\label{fig:free}
\end{figure}

\begin{figure}[hbt]
\centering
\vspace{0.3cm}
\graphicspath{{./figures/}}
\includegraphics[width=2.5in]{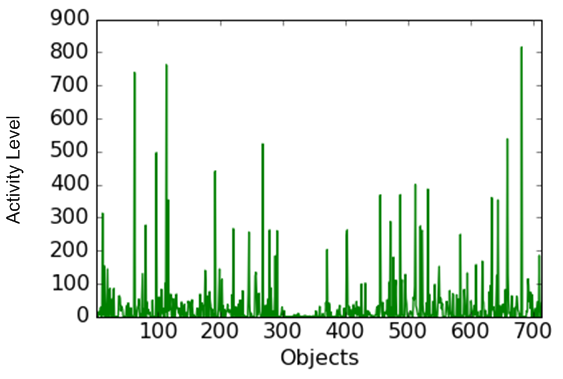}
\caption{Anomaly-free data (second dataset)}
\label{fig:free2}
\end{figure}

After detection and verification of anomalies by ground truth data and machine learning algorithm, we clean the data from such anomalous and abnormal activities making data anomaly-free. For preparation of anomaly-free data, we replace the anomalous activities of the users by average activities of all the users. Figure \ref{fig:free} and Figure \ref{fig:free2} show the anomaly-free data for first and second dataset respectively. It is clear from the plot of anomaly-free data that the users' activities are in a regular pattern. We train neural network model with anomaly as well as anomaly free data and observe mean square errors.

\subsection{Mean Square Error Evaluation}
\begin{figure}[h]
\centering
\begin{tabular}{ll}
\graphicspath{{./figures/}}
\includegraphics[scale=0.25]{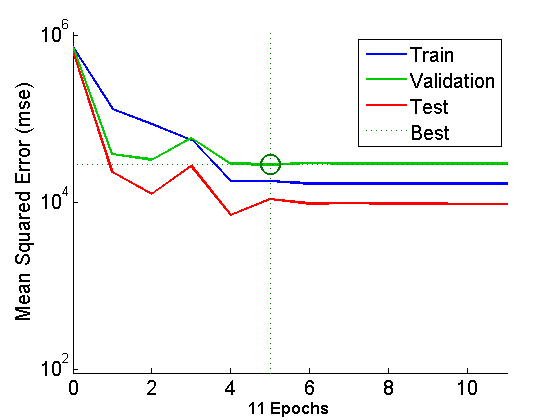}
&
\graphicspath{{./figures/}}
\includegraphics[scale=0.25]{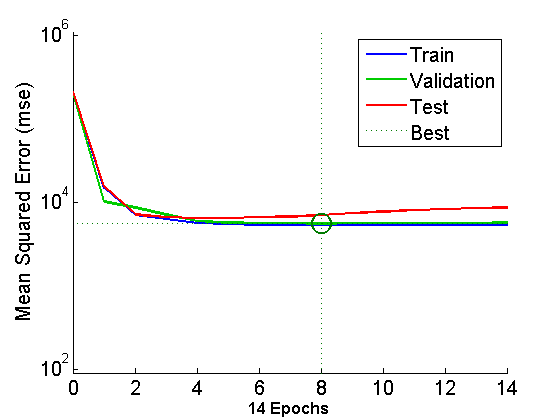}
\end{tabular}
\caption{Left: MSE of training model with anomalous data. 
Right: MSE of training model with anomaly-free data.}
\label{Fig:MSE}
\end{figure}

\begin{figure}[h]
\begin{tabular}{ll}
\graphicspath{{./figures/}}
\includegraphics[scale=0.25]{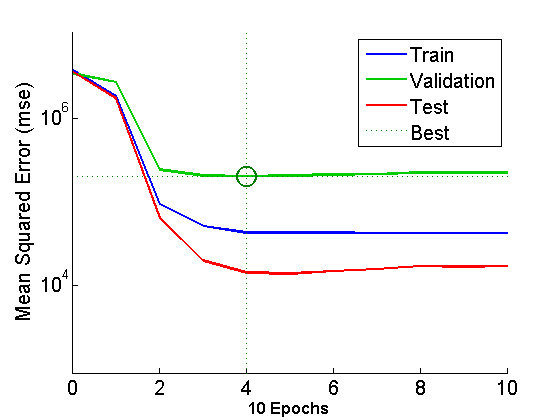}
&
\graphicspath{{./figures/}}
\includegraphics[scale=0.25]{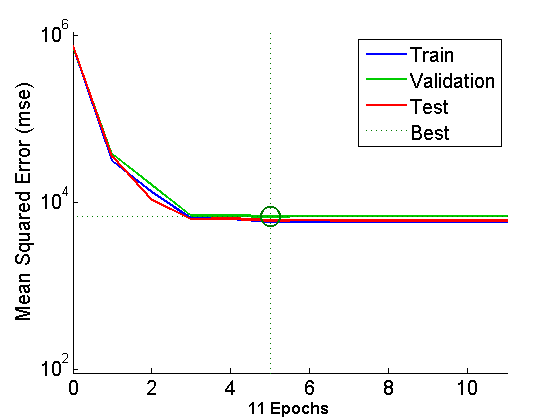}
\end{tabular}
\caption{Left: MSE of training model with anomalous data (second dataset). 
Right: MSE of training model with anomaly-free data (second dataset)}
\label{Fig:MSE2}
\end{figure}

For observing error difference in anomaly and anomaly-free data, we train a neural network model. We pass anomalous and anomaly-free data through this model and observe mean square error. Figure \ref{Fig:MSE} shows the mean square error for anomalous and anomaly-free data. It is observed that the mean square error of test, train and validation data is high when anomalous data is passed through the model. On the other hand, when the model is trained with anomaly free data, the overall mean square error is decreased. The mean square error of anomalous and anomaly-free data of second dataset is shown in Figure \ref{Fig:MSE2}. In order to highlight the significance of the pre-processing step for data pre-preparation, it is important to use a numeric metric. We calculate the mean square error for the model training with both anomalous and anomaly-free data. Thus, the mean square error serves as a numeric parameter to ascertain the impact of outliers in the cellular data. The mean square error calculation for both the datasets shows that anomaly free data help us develop a better model (as shown in Figure \ref{Fig:MSE} and Figure \ref{Fig:MSE2}.

The effect of the mean square error can be severe, depending upon the target of the model. For example in the case of sleeping cell detection, if anomalous data is used in the model, then the model would not be able to detect sleeping cells correctly. Under worst scenarios, this may lead to network outage as a consequence of denial of service for newer devices.

%

%

\section{ARIMA Time-series Forecasting Model}
\label{sec:ARIMA}
%
%
%
%
The CDR Datasets which are used in this work are time series data. CDR time series data can be used for predicting and detecting future anomalous behavior of the network and subscribers \cite{brockwell2016introduction, jason}. 
Among the time series forecasting models, ARIMA is a popular and widely used time-series forecasting model. ARIMA stands for \emph{Autoregressive Integrated Moving Average}. It is generalized auto-regressive model and adds the notion of integration \cite{arima}. The keys features of the model are explained below,

AR (Autoregression): This feature uses the dependency relationship between current observations and a specified number of previous observations. The mathematical representation of AR(p) is shown in Equation \ref{eq:1}. The mathematical representation of auto-regression model in equation \ref{eq:1} shows that the variable of interest \(Y_t\) is predicted by using the linear combination of its past values (\(Y_{t-1}\)):
\begin{align}
\label{eq:1}
{{\mathbf{Y_t}}} &= c+{\sum_{i=1}^{p} \varphi_i}{Y_{t-1}}+{\epsilon_t}                       
\end{align} 

where \(\varphi_i, . . . ,\varphi_p\) are parameters, c is constant and \(\epsilon_t\) is white noise.

I (Integrated): This feature is used for making a series stationary if it is non-stationary, done by subtracting raw observations. Mathematically, this feature is represented in the form of Equation \ref{eq:2}.
\begin{align}
\label{eq:2}
{{\mathbf{Y^`_t}}} &= {\mathbf{ Y_t}}-{Y_{t-1}}                 
\end{align} 
Where the consecutive observations are differences i.e., the past value of the variable (\(Y_{t-1}\)) is subtracted from the current value of (\(Y_t\)).

MA (Moving average): This feature uses dependency between an observation and a residual error from a moving average model applied to lagged observations. Equation \ref{eq:3} represents MA feature.
\begin{align}
\label{eq:3}
{{\mathbf{Y_t}}} &= {\mathbf{\mu}}+{\epsilon_t}+{\sum_{i=1}^{q} \theta _i}{\epsilon_{t-1}}                       
\end{align}
where \(\theta_i, . . . ,\theta_q\) are parameters, \(\mu\) is expectation of \(Y_t\) and \(\epsilon_t\) , \(\epsilon_{t-1}\) are white noise error terms.

\begin{figure}[ht!]
\centering
\vspace{0.2cm}
\graphicspath{{./figures/}}
\includegraphics[scale=0.3]{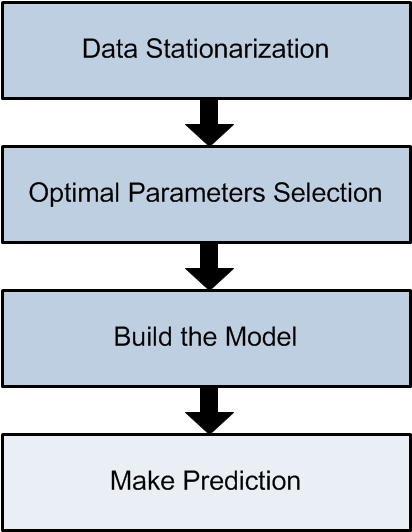}
\caption{Time series process}
\label{fig:timeseries}
\end{figure}

ARIMA time series forecasting model works in following steps as shown in Figure \ref{fig:timeseries}.

\subsection{Data Stationarity}
\begin{figure}[h]
\centering
\vspace{0.3cm}
\graphicspath{{./figures/}}
\includegraphics[width=2.5in]{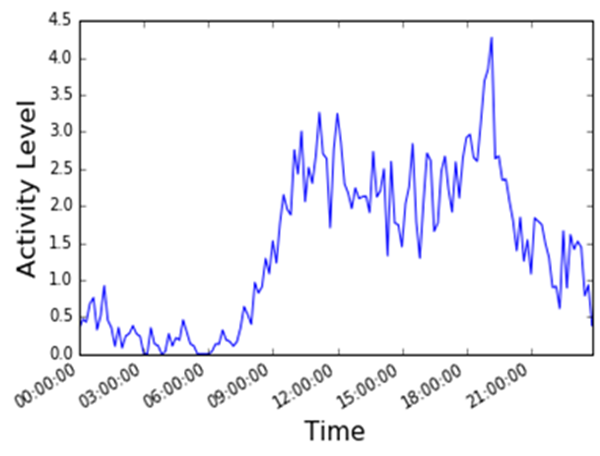}
\caption{Time series data plot}
\label{fig:timeseries_plot}
\end{figure}

We visualize the data and observe increasing or decreasing trends of the data. This time series data is non-stationary, as can be observed from Figure \ref{fig:timeseries_plot}. We take necessary action to make the data stationary as required for time series forecasting model.  
%

It is typically assumed (and required) for time series forecasting models that the input data is stationary. If time series data is not stationary, it should be made stationary before training a time series forecasting model\footnote{For determining the stationarity of time series models, there are different methods such as visualizing the time series's plot, observing summary statistics, augmented Dickey-Fuller test.}.
In summary statistics, it is observed that for stationary time series data, mean and variance should be constant for observations.

\begin{figure}[h]
\centering
\vspace{0.3cm}
\graphicspath{{./figures/}}
\includegraphics[width=2.5in]{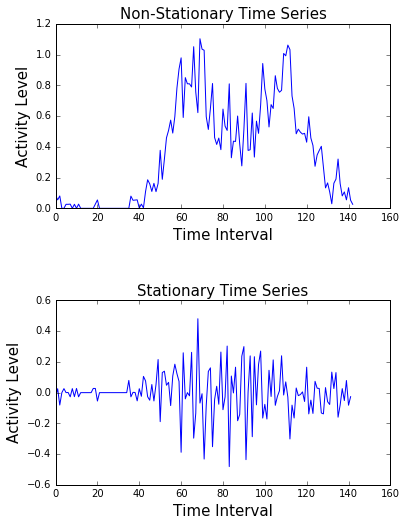}
\caption{Non-stationary and stationary time series. 
(As CDR activities of users are aggregated on 10 minutes interval in available dataset, so x-axis represents the 10 minutes intervals of one day's CDR activity and y-axis represents activity level)}
\label{fig:timeseriesCOMPARISON}
\end{figure}

For stationarity confirmation of the data, The Augmented Dicky-Fuller (ADF) statistical test is used. ADF test is also called unit root test. The ADF test has two hypothesis; \emph{Null hypothesis and Alternate hypothesis}. Null hypothesis suggests that time series data has a unit root, so data is non-stationary. Alternate hypothesis suggests that time series data does not have any unit root implying that data is stationary. These hypothesis are interpreted by \textit{p} value of the ADF test. If \textit{p}-value is greater than 0.05, null hypotheses is accepted and data is non-stationary. Similarly, if \textit{p}-value is less than or equal to 0.05, null hypothesis is rejected and data is stationary. We have applied ADF test on one day's CDR activities of the users for confirmation of data stationarization. ADF statistics of available dataset shows that \textit{p}-value is 0.6177, implying that the data is non-stationary. 
Hence after confirmation of non-stationary behavior of data, we make the data stationary by differencing. 
For comparison purpose, non-stationary and stationary time series are plotted together in Figure \ref{fig:timeseriesCOMPARISON}. The upper part of Figure \ref{fig:timeseriesCOMPARISON} represents the non-stationary time series and the lower part represents the stationary time series.

\subsection{Parameter Estimation}
After determining and confirming the stationarity of the time series, the next step is to determine the optimal value of model's parameters AR(p) and MA(q). The optimal values of p and q is determined with the help of auto-correlation function (ACF) and partial auto-correlation function (PACF) plots.
\subsubsection{Autocorrelation Function (ACF)}
The plot of ACF in Figure \ref{fig:ACF_PACF} shows the correlation between an observations with lag values. On the x-axis of the ACF plot is number of lags and on y-axis is the correlation coefficient. The model is AR if the ACF trails off after a lag and has a hard cut-off in the PACF after a lag. This lag is taken as the value for p.
\subsubsection{Partial Autocorrelation Function (PACF)}
\begin{figure}[h!]
\centering
\vspace{0.5cm}
\graphicspath{{./figures/}}
\includegraphics[width=2.5in]{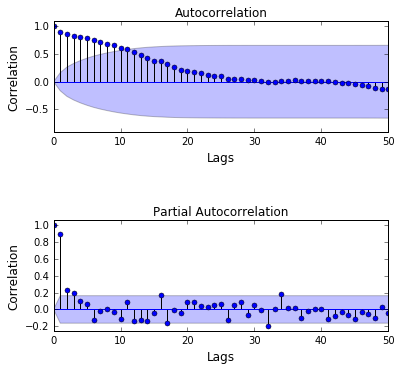}
\caption{ACF and PACF plot of time series data}
\label{fig:ACF_PACF}
\end{figure}

The plot summarizes the correlations for an observation with lag values not accounted for by prior lagged observations.
The model is MA if the PACF trails off after a lag and has a hard cut-off in the ACF after the lag. This lag value is taken as the value for q.
The model is a mix of AR and MA if both the ACF and PACF trail off. The plots of ACF and PACF for one day's CDR activity of the users are shown in Figure \ref{fig:ACF_PACF}. The cone in the plot represents the confidence interval which shows that the correlations outside this cone are very likely a correlation. In the plot of ACF and PACF, all lag values are printed which makes plot difficult to visualize, for ease of visualization of plot we limit the lag values to 50. After visualizing the plots of ACF and PACF, the order of AR and MA is decided.

\subsection{Build the ARIMA model}
After defining the optimal values and the order of model's parameters $(p,d,q)$, we build ARIMA model and fit data in the model. The dataset is divided into train and test sequences. The model is trained on 70 percent data and tested on remaining 30 percent data.
\subsection{Make Prediction}
\begin{figure}[h!]
\centering
\vspace{0.3cm}
\graphicspath{{./figures/}}
\includegraphics[width=2.5in]{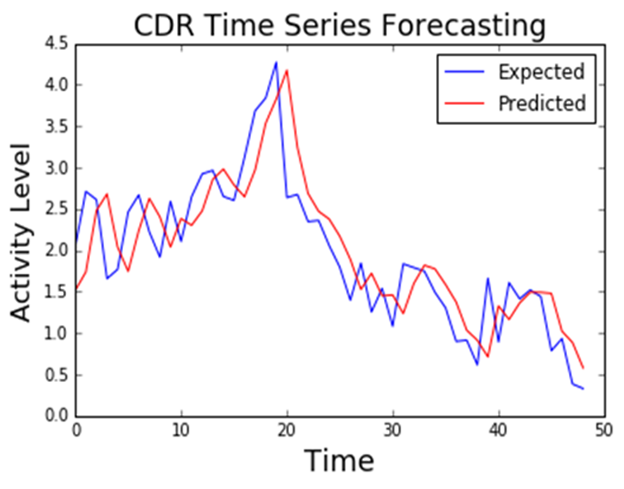}
\caption{Time series process forecasting}
\label{fig:timeseries_forecasting}
\end{figure}

\begin{figure}[h!]
\centering
\vspace{0.3cm}
\graphicspath{{./figures/}}
\includegraphics[width=2.5in]{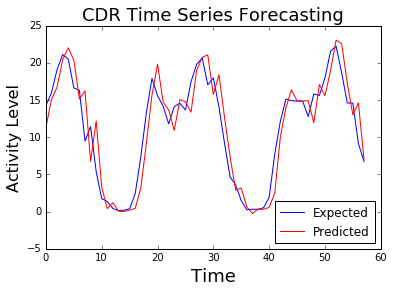}
\caption{Time series process forecasting over a week}
\label{fig:timeseries_forecasting_week}
\end{figure}

After building the model, we use it for forecasting. Firstly we analyze the model with the available data, in this model precision is observed. Figure \ref{fig:timeseries_forecasting} shows prediction with ARIMA model applied on data for one day. We also apply the model on user's activity of a week. Figure \ref{fig:timeseries_forecasting_week} shows the prediction of user's activity over a week.

With such time series forecasting for CDR data, the trends in mobile network as well as mobile's subscribers are predicted well before time. This help in interpretation, and thus smart management of cellular networks in terms of spectrum management, fault detection/avoidance and provision of just-in-time services. These applications have great potential use in next generation networks. 

\section{Conclusion and Future Direction}
\label{sec:Conclusion}
In this work, we analyzed CDR data from mobile network. For CDR data analysis, we used k-means clustering technique. The users' activities which exhibit unusual behavior are termed as anomalies. We verified anomalies by plotting ground truth data and analysis plot for k-means clustering algorithm. After detection and verification of anomalies, we can also identify the region where such anomalies occur. This helps in identification of region of interest (an important geo location) or event of interest (an extraordinary incident). After identification of such regions or events, proper action such as resource distribution, sending drone small cells etc can be taken in advance and on time. Hence because of such actions, the users requirements will be fulfilled and will have best QoS as well as network congestion will be avoided.

CDR data analytics will help in understanding the dynamics of the users which will helpful in empowering the concept of smart colonies. Also the hidden information obtained from CDR data analytics of user will be helpful in efficient resource allocation and distribution.

\ifCLASSOPTIONcaptionsoff
  \newpage
\fi



%
%
\bibliographystyle{IEEEtran} 
\bibliography{mybib} 

%
%

%
\begin{IEEEbiography}[{\includegraphics[width=1in,height=1.25in,clip,keepaspectratio]{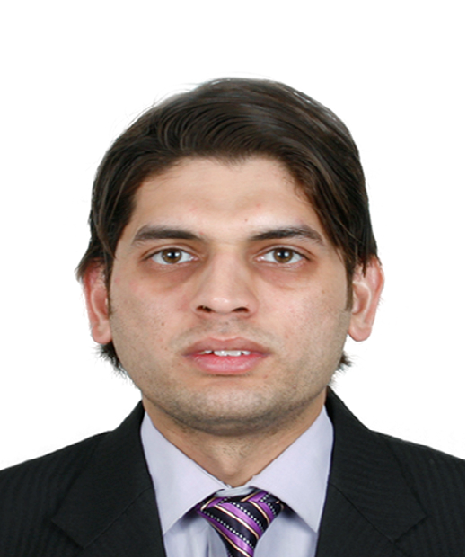}}]{Kashif Sultan}
received Masters degree in Information and Communication Engineering from University of Science and Technology Beijing, China, in 2015. Currently, he is a PhD researcher at the School of Computer and Communication Engineering, University of Science and Technology Beijing, China. He is the recipient of USTB Chancellors scholarship. His research interests include communication signal processing, next generation networks and big data analytics.
\end{IEEEbiography}

\begin{IEEEbiography}[{\includegraphics[width=1in,height=1.25in,clip,keepaspectratio]{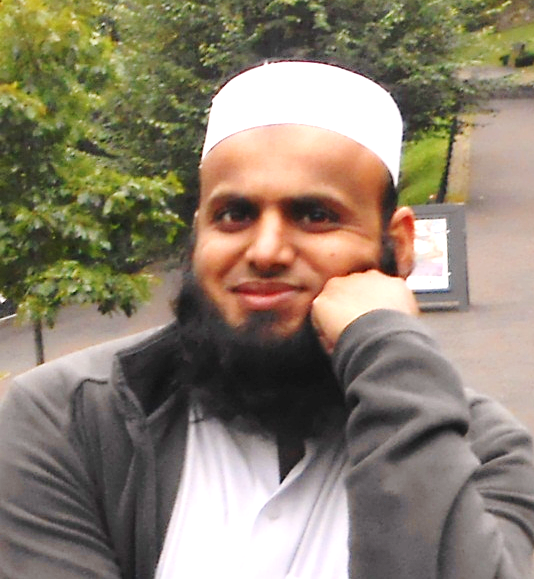}}]{Hazrat Ali}
received in BSc and MSc degrees in electrical engineering in 2009 and 2012 respectively. He did his PhD in 2015 from University of Science and Technology Beijing, China. He is currently Assistant Professor at Department of Electrical Engineering, COMSATS Institute of Information Technology Abbottabad. At CIIT, he is the member of the signal processing and machine learning research group. He is also the course head for Digital Signal Processing course. His research interests lie in unsupervised learning, generative and discriminative approaches, and speech and image processing. 
He is Associate Editor at IEEE and served as reviewer at IEEE Access, IEEE Transactions on Neural Networks and Learning Systems, Springer Neural Processing Letters, ACM Transactions on Asian and Low Resource Language Information Processing, Elsevier Computers and Electrical Engineering, International Journal of Artificial Intelligence Tools, Journal of Experimental and Theoretical Artificial Intelligence, Transactions on Internet and Information Systems, Springer Multimedia Tools and Applications and as PC member at Frontiers of Information Technology conference (FIT 2016), ICACT 2018, ICACT 2017, and IEEE WiSPNet 2018. He is selected as young researcher at the 5th Heidelberg Laureate Forum, Heidelberg, Germany. He is the recipient of the HEC Scholarship, IEEE Student Travel Award, the IBRO grant, the TERENA/CISCO Travel grant, QCRI/Boeing Travel grant and the Erasmus Mundus STRoNGTiES research grant.
\end{IEEEbiography}


\begin{IEEEbiography}[{\includegraphics[width=1in,height=1.25in,clip,keepaspectratio]{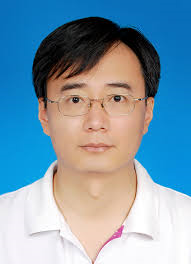}}]{Zhongshan Zhang}
 received the B.E. and M.S. degrees in computer science from the Beijing University of Post and Telecommunications (BUPT) in 1998 and 2001, respectively, and the Ph.D. degree in electrical engineering from BUPT in 2004. From 2004, he joined the DoCoMo Beijing Laboratories as an Associate Researcher, and was promoted to be a Researcher in 2005. In 2006, he joined the University of Alberta, Edmonton, AB, Canada, as a Post-Doctoral Fellow. In 2009, he joined the Department of Research and Innovation, Alcatel-Lucent, Shanghai, as a Research Scientist. From 2010 to 2011, he was with the NEC China Laboratories, as a Senior Researcher. He is currently a Professor with the School of Computer and Communication Engineering, University of Science and Technology Beijing. His main research interests include statistical signal processing, self-organized networking, cognitive radio, and cooperative communications. He served or is serving as a Guest Editor and/or an Editor for several technical journals, such as the IEEE Communications Magazine and the KSII Transactions on Internet and Information Systems.
\end{IEEEbiography}




\end{document}